\pdfoutput=1
\documentclass[11pt]{article}

\usepackage{ACL2023}

\usepackage{times}
\usepackage{latexsym}
\usepackage[T1]{fontenc}
\usepackage[utf8]{inputenc}
\usepackage{microtype}
\usepackage{inconsolata}

\newcommand{\tvd}{$\textsc{TV}_d$}
\newcommand{\tvds}{$\textsc{TV}_d$ }
\newcommand{\q}[1]{``#1''}

\usepackage{amsmath}
\usepackage{graphicx}
\usepackage{listings}
\usepackage{tabularx}

\title{Temporal Validity Change Prediction}

\author{Georg Wenzel \\
  University of Innsbruck, Austria \\
  \texttt{georg.wenzel@outlook.com} 
  \\\And
  Adam Jatowt \\
  University of Innsbruck, Austria \\
  \texttt{adam.jatowt@uibk.ac.at} \\}

\begin{document}
\maketitle
\begin{abstract}
Temporal validity is an important property of text that is useful for many downstream applications, such as recommender systems, conversational AI, or story understanding. Existing benchmarking tasks often require models to identify the temporal validity duration of a single statement. However, in many cases, additional contextual information, such as sentences in a story or posts on a social media profile, can be collected from the available text stream. This contextual information may greatly alter the duration for which a statement is expected to be valid. We propose \emph{Temporal Validity Change Prediction}, a natural language processing task benchmarking the capability of machine learning models to detect contextual statements that induce such change. We create a dataset consisting of temporal target statements sourced from Twitter and crowdsource sample context statements. We then benchmark a set of transformer-based language models on our dataset. Finally, we experiment with temporal validity duration prediction as an auxiliary task to improve the performance of the state-of-the-art model.
\end{abstract}

\section{Introduction}
\label{sec:introduction}
In human communication, temporal properties are frequently underspecified when authors assume that the recipient can infer them via commonsense reasoning. For example, when reading \q{I am moving on Saturday}, a reader is likely to assume the person will be busy for most of the day. On the other hand, when reading \q{I will make a sandwich on Sunday}, this is likely to only take up a fraction of the author's day and may not impact other plans. Such reasoning is referred to as \emph{temporal commonsense} (TCS) \emph{reasoning} \citep{tcssurvey}. 

\emph{Temporal validity} \citep{content-expiry-date,tnli,cta} is a property that is vital for the proper understanding of a text. The temporal validity of a statement, i.e., whether it contains valid information at a given time, often requires TCS reasoning to resolve. For example, in determining whether a statement like \q{I am driving home from work} is still valid after five hours, we may use our prior understanding of the typical duration of related events, such as commuter traffic. While the amount of research into TCS and, to a degree, temporal validity, has risen over the past years \citep{tcssurvey}, there are still several properties of temporal validity that have not been considered in previous research. One such property is the impact of \emph{context} on the temporal validity duration of a statement. For example, the sentence \q{I am driving home from work} may be valid for a longer time when followed by a statement such as \q{There is a massive traffic jam}. 

\begin{figure}[tbp]
    \centering
    \includegraphics[width=\columnwidth]{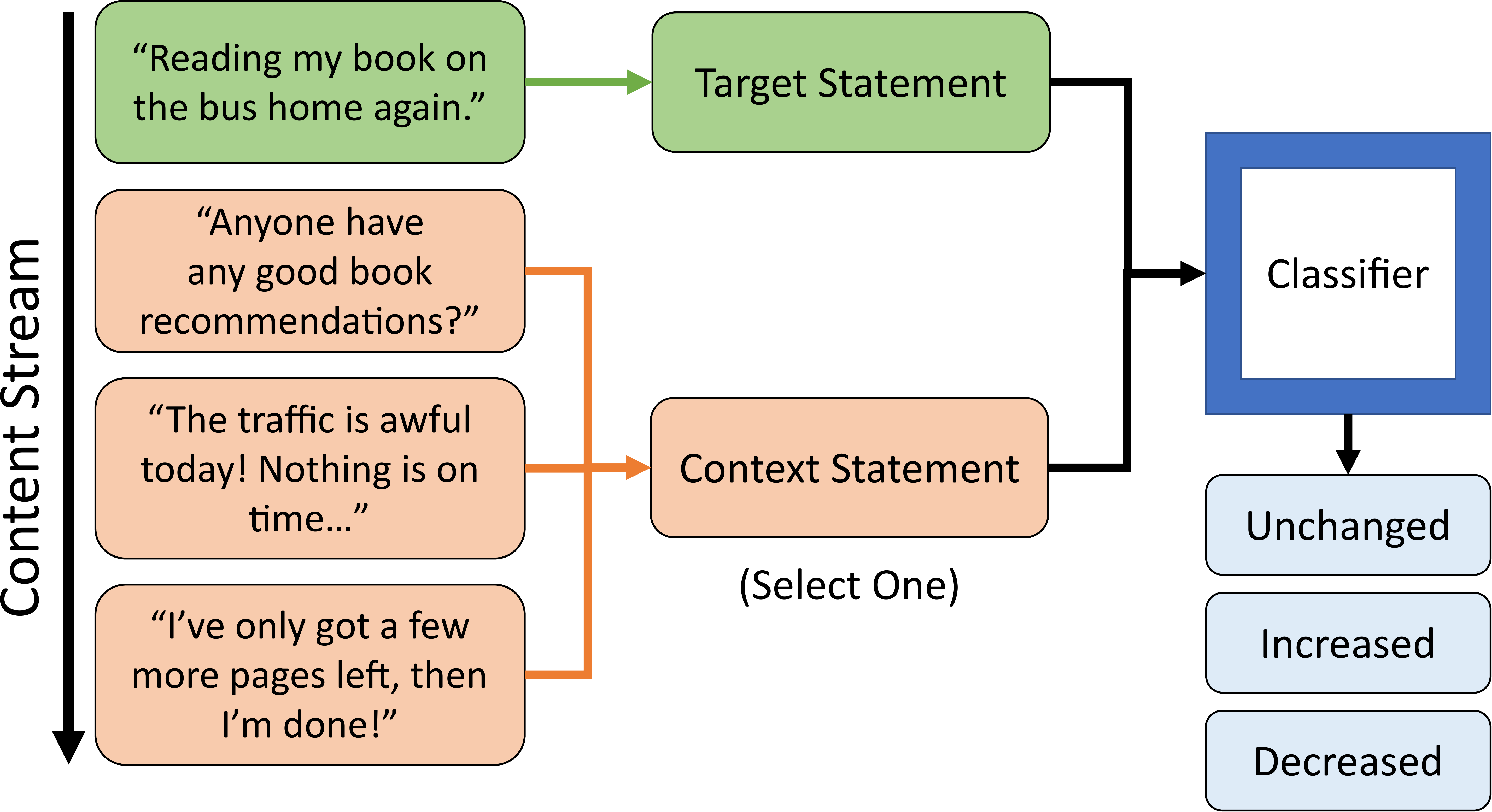}
    \caption{A visualization of the \textsc{TVCP} task}
    \label{fig:tvcptask}
\end{figure}

To model this problem, we propose a new NLP task format called \emph{Temporal Validity Change Prediction} (\textsc{TVCP}), which requires reasoning over whether a context statement changes the temporal validity duration of a target statement. The task is visualized in Figure \ref{fig:tvcptask}. We propose the following applications for such a system.

\textbf{Timeline Prioritization:} Social media services such as Twitter rely on recommender systems to prioritize the vast amount of content that their users produce. One possible way to improve the prioritization of content is to consider its temporal validity \citep{tweetclassification,tweetlifetime}, as users are likely to care about current and relevant information over more general, stationary statements. \textsc{TVCP} can be used to leverage the stream of social media posts by a given user as possible context to better estimate the temporal validity duration of a previously observed post.
    
\textbf{User Status Tracking:} Similarly, the content of a user's posts on social media could be utilized for other analytical or business purposes, such as predicting revenue streams \citep{tweetprediction-rev,tweetprediction-stock,tweetprediction-iphone,tweetprediction-rev2} or identifying trends in a community's or an individual user's behaviour \citep{tweetprediction-location,tweetprediction-life,tweetprediction-action}. \textsc{TVCP} could be used to identify posts that refer to previous temporal information, to detect chains of thought about topics that may not be self-contained.

\textbf{Conversational AI:} Foundation models, such as \textsc{ChatGPT} \citep{instructgpt} and \textsc{Bard} \citep{bard}, could use the temporal validity of statements provided by the user to keep track of knowledge that is still relevant to the conversation. Using \textsc{TVCP}, new messages by the user could be evaluated to adjust the expected temporal validity period of previously learned facts. This is especially relevant as initial reports indicate that foundation models may struggle with TCS reasoning \citep{chatgpt-tcs}.

Our main contributions are the following:

\begin{enumerate}
    \item We define a novel NLP task (\textsc{TVCP}). This ternary classification task requires models to predict the impact of a context statement on a target statement's temporal validity duration.
    
    \item We build a dataset of tuples consisting of time-sensitive \emph{target statements}, as well as \emph{follow-up statements} that act as context for our task.
    
    \item We evaluate the performance of existing pre-trained \emph{language models} (LMs) on our dataset, including models fine-tuned on other TCS tasks as well as \textsc{ChatGPT}. 
    
    \item We propose an augmentation to the training process that leverages temporal validity duration information to help improve the performance of the state-of-the-art classifier.
\end{enumerate}

\section{Related Work}
\subsection{Temporal Commonsense Reasoning}
TCS reasoning is often considered one of several categories of commonsense reasoning \citep{commonsense-survey-2,commonsense-survey-1}. A major driver of research specifically into TCS appears to have been the transformer architecture \citep{transformer} and resulting LMs. In recent years, several datasets that specifically aim to benchmark TCS understanding have been published \citep{mctaco,torque,wikihowstepordering,timedial,tracie}, while \textsc{ROCStories} \citep{rocstories} appears to be the only dataset focussing on this type of reasoning before the publication of the transformer architecture. Small adjustments to transformer-based LMs are often proposed as state-of-the-art solutions for these datasets \citep{alice,timeawarept,taco-lm,alicepp,ensemble-bert-tcs,tracie,eventoptimaltransport,sleer,cocolm}. Similarly, temporalized transformer models are popular solutions for tasks such as document dating or semantic change detection \citep{tempattention,tempobert,timebert}.

The TCS taxonomy defined by \citet{mctaco} is frequently referenced. It contains the five dimensions of \emph{duration} (how long an event takes), \emph{temporal ordering} (typical order of events), \emph{typical time} (when an event happens), \emph{frequency} (how often an event occurs) and \emph{stationarity} (whether a state holds for a very long time or indefinitely).

\subsection{Temporal Validity}
\begin{table*}[htbp]
    \small
    \centering
    \begin{tabular}{|l|l|l|l|l|r|}
        \hline
         Method & Task & Data Source  & Duration Bias & Model & \# Samples\\
         \hline
         \citet{tweetclassification} & \tvd & Twitter & N/A & SVC & 9,890\\
         \citet{content-expiry-date} & \tvd & Blogs, News, Wikipedia & \emph{years} & SVC & 1,762\\
         \citet{tnli} & \textsc{TNLI} & Image Captions & \emph{seconds}\footnotemark & LM & 10,659\\
         \citet{cta} & \tvd & WikiHow & \emph{hours} & LM & 339,184\\
         Ours & \textsc{TVCP} & Twitter & \emph{hours} & LM & 5,055\\
         \hline
    \end{tabular}
    \caption{Summary of related work}
    \label{tab:related}
\end{table*}

Compared to TCS reasoning, temporal validity in text is a less well-researched field. It effectively combines three dimensions of the taxonomy by \citet{mctaco}: \emph{Stationarity}, to reason about whether a statement contains temporal information, \emph{typical time}, to reason about when the temporal information occurs, and \emph{duration}, to reason about how long the temporal information takes to resolve. 

\citet{tweetclassification} classify the lifetime duration of tweets, i.e., the informational value of a tweet over time. They use handcrafted, domain-specific features to train a \emph{support vector classifier} (SVC) on supervised samples. 

\citet{content-expiry-date} similarly design features to estimate the temporal validity duration of sentences collected from news, blog posts, and Wikipedia using SVCs. Their features contain general properties such as the word- or sentence length, but also more complex ones, such as latent semantic analysis.

\citet{tnli} define the \emph{Temporal Natural Language Inference} (\textsc{TNLI}) task. The goal of \textsc{TNLI} is to determine whether the temporal validity of a given hypothesis sentence is supported by a premise sentence.

\citet{cta} build a large dataset of human annotations specifying the duration required to perform various actions on WikiHow as well as their respective temporal validity durations. 

\subsection{Comparison with Related Work}
\footnotetext[1]{Based on analysis of a sample. \tvds labels are not available for the full dataset.}
Table \ref{tab:related} shows the most closely related research. As noted, our dataset is based on the proposed \textsc{TVCP} task, whereas previous work was based on the \tvds and \textsc{TNLI} tasks. All tasks are described in more detail in Section \ref{sec:task}.

Another prominent distinctive attribute is the text source and the resulting temporal validity duration bias. For example, sentences sourced from news or Wikipedia articles often appear to be valid for years or longer. On the other hand, image captions may only be valid for a few seconds or minutes. We decided to source our sentences from Twitter due to its alignment with our downstream use cases. Similar to \citet{cta}, our collected temporal information tends to be valid for a few hours.

We follow recent research by evaluating our dataset using transformer-based LMs, whereas earlier approaches relied on methods such as SVCs. 

Except for the \textsc{CoTAK} dataset \citep{cta}, the datasets tend to be relatively small. As crowdsourcing is used in all datasets referenced in Table \ref{tab:related} to annotate text spans with commonsense information, the costs of dataset creation can quickly escalate. In addition, we ask participants to create examples of follow-up statements that cause temporal validity change. This approach further restricts the overall size of our dataset due to the relative difficulty of the task. 

\section{Task}
\label{sec:task}
\subsection{Defining Temporal Validity}
Temporal validity, in essence, is simply the time-dependent validity of a text. As shown in Equation \ref{eq:tv}, the temporal validity of a statement $s$ at a time $t$ is a binary value that determines whether the information in $s$ is valid at the given time.

\begin{equation}
\label{eq:tv}
\small
\textsc{TV}(s, t) = 
\begin{cases} 
\text{True} & \text{if information in $s$ is valid at $t$}, \\
\text{False} & \text{otherwise}
\end{cases}
\end{equation}

\begin{figure}[htbp]
    \centering
    \includegraphics[width=0.8\columnwidth]{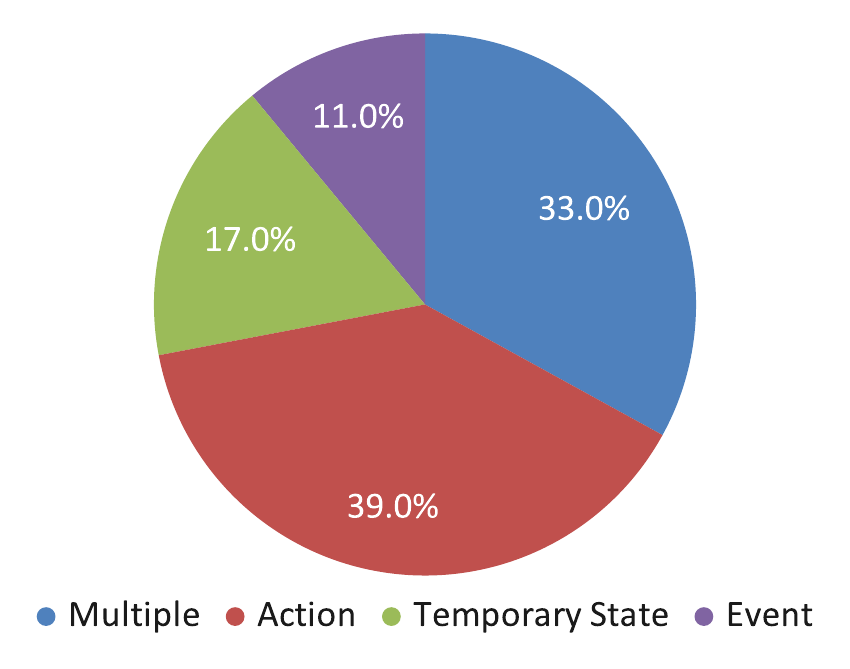}
    \caption{Distribution of different types of temporal information in a sample of our dataset}
    \label{fig:information-types}
\end{figure}

In some previous research \citep{tnli,cta}, the scope of evaluated temporal information is limited to actions, such as \q{I am \emph{baking bread}}. However, we note that other types of temporal information exist, such as events (e.g., in the sentence \q{\emph{Job interview} tomorrow}) or temporary states (e.g., in the sentence \q{It \emph{is nice out} today}). In an analysis of a subset of our collected statements, shown in Figure \ref{fig:information-types}, we find that these alternative types of temporal information constitute a significant portion (28\%) of samples. Additionally, one-third of sampled statements contained at least two distinct pieces of temporal information with differing temporal validity spans. This indicates that the true scope of determining the temporal validity of a text may exceed what current datasets are benchmarking.

We assume that the temporal validity of stationary statements is constant for any given timestamp $t$. A stationary statement may be continuously true (e.g., \q{Japan lies in Asia}), or continuously false (e.g., \q{Japan lies in Europe}). This includes information that is fully contained in the past (e.g., \q{I went to the bank yesterday}). In general, we do not expect the validity of such a statement to change. 

For contemporary or future information, we assume the statement is valid from the moment of sentence conception until the information is no longer ongoing. We include the duration of the information, rather than just its occurrence time, as humans are likely to still consider durative information relevant while it is ongoing. For example, we may reason that the statement \q{I will take a shower at 8 p.m.} still has informational value at 8:05 p.m., as it allows us to infer the current action of the author. 

\subsection{Formalizing Existing Tasks}
\subsubsection{Temporal Validity Duration Estimation}
\emph{Temporal Validity Duration Estimation} (\tvd) is the primary task that is evaluated in temporal validity research \citep{tweetclassification,content-expiry-date,cta}. The goal is to estimate the duration for which the statement is valid,  starting at the statement creation time. We formalize this task in Equation \ref{eq:tvd}, where $t_s$ is the timestamp at which the statement $s$ is created.

\begin{equation}
\label{eq:tvd}
\small
\textsc{TV}_d(s) = \max_{t \geq t_s} \{t \,|\, \textsc{TV}(s, t) = \text{True}\}
\end{equation}

The \tvds task is useful in downstream applications such as social media, where information on the posting time of a statement is readily available and can be used to infer the span during which the statement is valid.

\subsubsection{Temporal Natural Language Inference}

The goal of \textsc{TNLI} \citep{tnli} is to infer whether a statement is temporally valid, given additional context, using typical NLI terminology \citep{nlibook,nlisurvey}. \textsc{TNLI} requires a \emph{hypothesis statement} (that we call \emph{target statement}, or $s_t$) and a \emph{premise sentence} (that we call \emph{follow-up statement}, or $s_f$). Implicitly, the inference takes place at $t_{s_f}$, that is, the posting time of the follow-up statement, but no explicit duration information is required to solve this task. Formally, we define \textsc{TNLI} in Equation \ref{eq:tnli} (\textsc{SUO} = supported, \textsc{INV} = invalidated, \textsc{UNK} = unknown), where $\textsc{TV}^{c}(s,t)$ is the temporal validity of a statement $s$ at a time $t$ given context $c$. The \textsc{UNK} class is assigned in cases where $\textsc{TV}^{s_f}(s_t, t_{s_f})$ is neither clearly supported nor invalidated by the context.

\begin{equation}
    \small
    \label{eq:tnli}
    \textsc{TNLI}(s_t, s_f) = 
    \begin{cases}
        \textsc{SUO} & \textsc{TV}^{s_f}(s_t, t_{s_f})=\text{True}\\
        \textsc{INV} & \textsc{TV}^{s_f}(s_t, t_{s_f})=\text{False}\\
        \textsc{UNK} & \textsc{TV}^{s_f}(s_t, t_{s_f})=\text{Unclear}
    \end{cases}
\end{equation}

Unlike \tvd, this task format lends itself to downstream applications such as story understanding, wherein a larger text stream of individual statements is provided with no clear explicit notion of time passing between each sentence (e.g., in a book).

\subsection{Temporal Validity Change Prediction}
We propose \emph{Temporal Validity Change Prediction} (\textsc{TVCP}), which combines ideas from both the inference- and duration-based tasks. Like \textsc{TNLI}, we require $s_t$ and $s_f$ for classification, and determine a ternary label that provides information about the impact of $s_f$ on $s_t$. Unlike \textsc{TNLI}, our goal is to predict a \emph{change} in the temporal validity \emph{duration} of $s_t$.

We consider \textsc{TVCP} a necessary step in accurately determining a statement's temporal validity. Simply estimating the duration of the statement alone may not yield very precise results when it is, as in many use cases, extracted from a rich context, such as a book, a story, a news article, a step-by-step guide, or a social media profile. In these cases, surrounding information may provide additional context that could lead us to a different \tvds estimate. Simply concatenating $s_t$ and $s_f$ may lead to the classification of temporal information within $s_f$, which is undesired. Our segmentation of $s_t$ and $s_f$ into different semantic roles, similar to \textsc{TNLI}, prevents this issue.

\begin{figure}[htbp]
    \centering
    \includegraphics[width=\columnwidth]{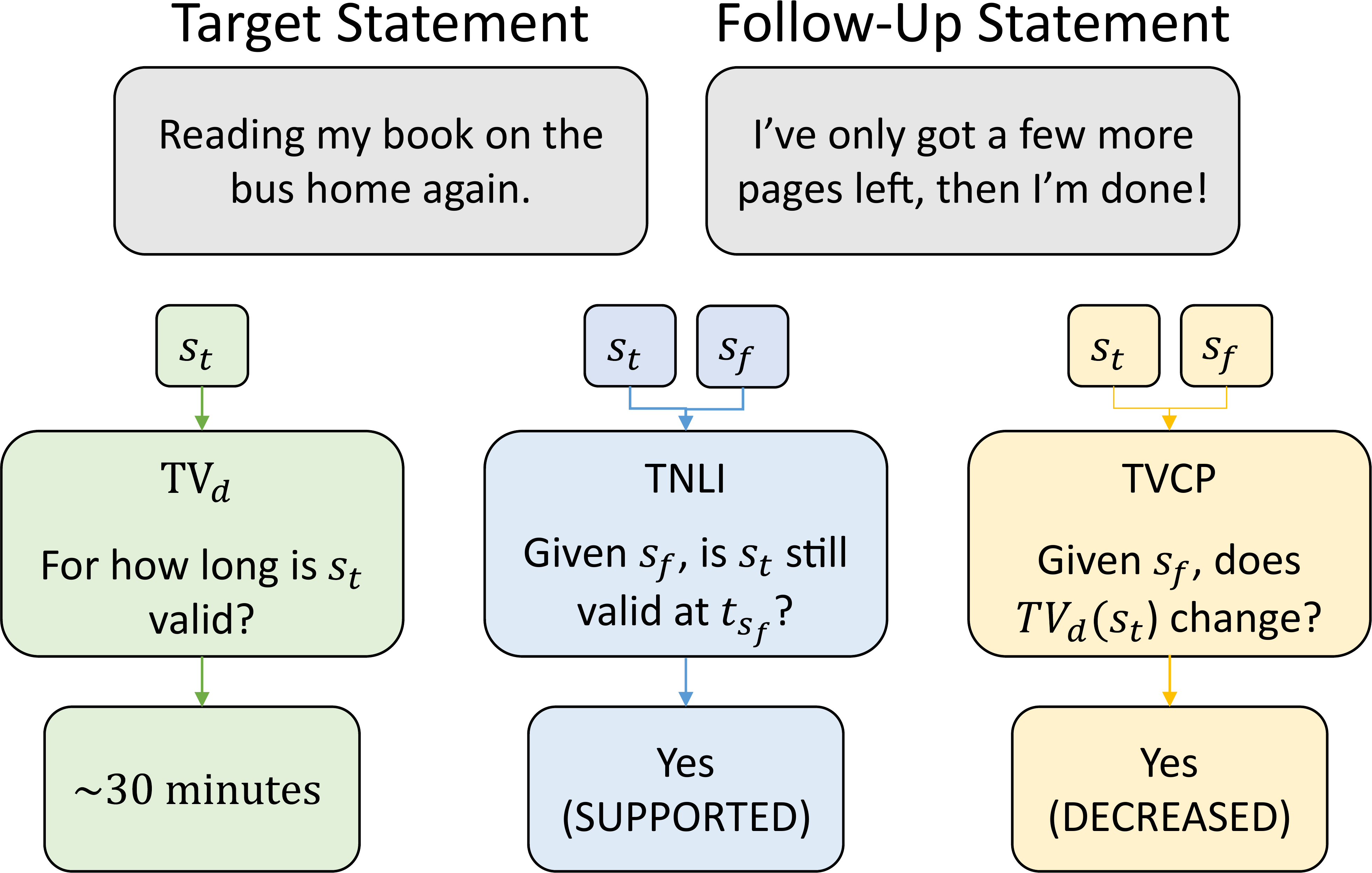}
    \caption{An example of \tvd, \textsc{TNLI} and \textsc{TVCP}}
    \label{fig:tasksummary}
\end{figure}

Formally, we define \textsc{TVCP} in Equation \ref{eq:tvcp} (\textsc{DEC} = decreased, \textsc{UNC} = unchanged, \textsc{INC} = increased), where $\textsc{TV}_d^{c}(s)$ is the temporal validity duration of a statement $s$ given context $c$. Figure \ref{fig:tasksummary} shows a concrete comparative example of the goal of all three tasks.

\begin{equation}
    \small
    \label{eq:tvcp}
    \textsc{TVCP}(s_t, s_f) = 
    \begin{cases}
        \textsc{DEC} & \textsc{TV}_d(s_t) > \textsc{TV}_d^{s_f}(s_t)\\
        \textsc{UNC} & \textsc{TV}_d(s_t) = \textsc{TV}_d^{s_f}(s_t)\\
        \textsc{INC} & \textsc{TV}_d(s_t) < \textsc{TV}_d^{s_f}(s_t)
    \end{cases}
\end{equation}

Since \textsc{TVCP} is a signal measuring the difference between \tvds with- and without $s_f$, respectively, a more fine-grained \tvds classification increases the number of \textsc{TVCP} instances that can be detected. On the other hand, evaluating \tvds on a very fine-grained scale may be more difficult for both models and humans \citep{roundnumberbias}, and the resulting uncertainty and inaccuracies could lead to a degradation of the system as a whole.

\begin{figure}
    \centering
    \includegraphics[width=0.7\columnwidth]{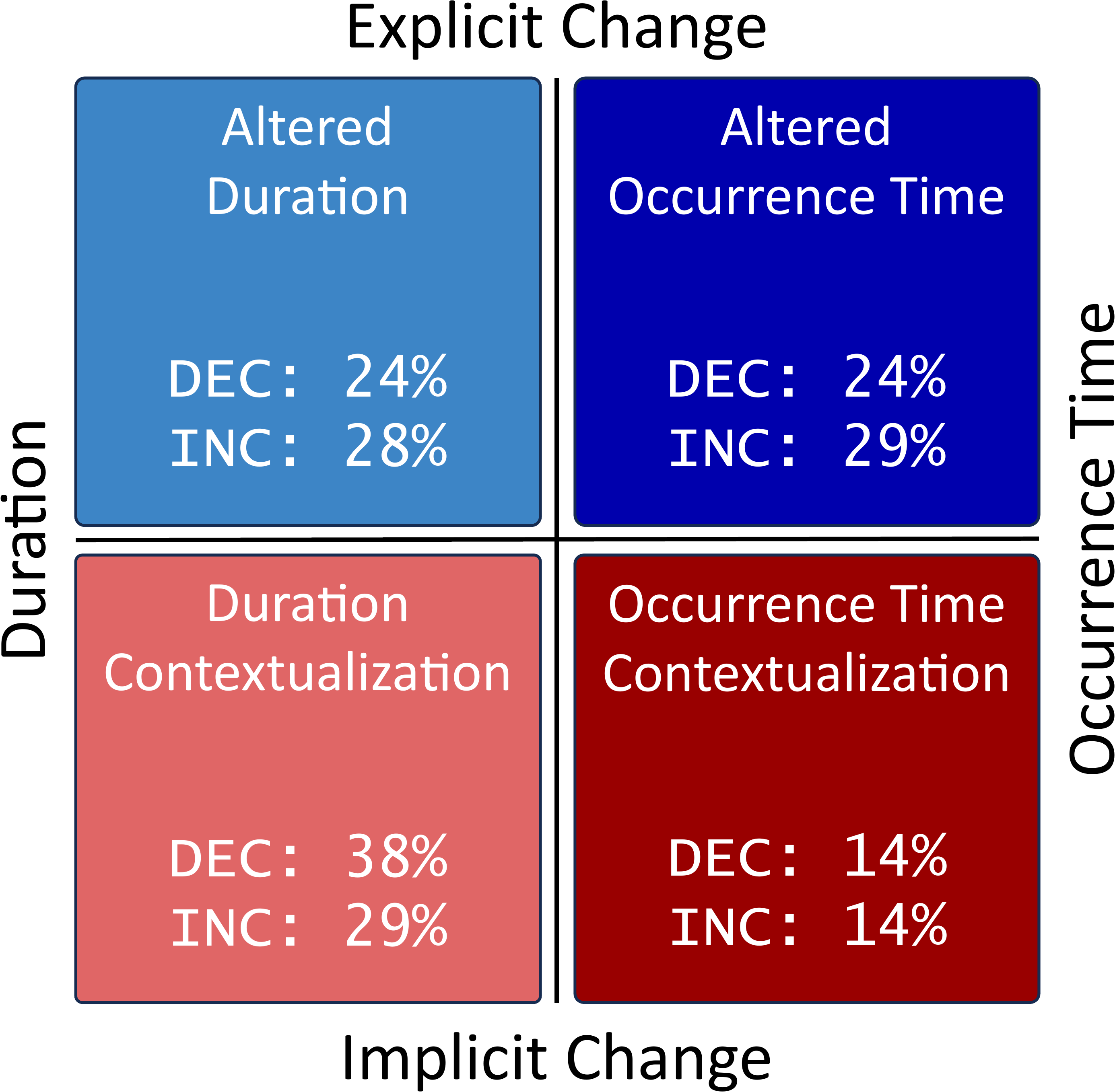}
    \caption{Dimensions of temporal validity change. The frequency of each category for \textsc{DEC} and \textsc{INC} classes in our sample is appended.}
    \label{fig:changedimensions}
\end{figure}

In our sample analysis, we find that temporal validity change generally occurs along two axes, shown in Figure \ref{fig:changedimensions}. The first dimension is \textbf{\emph{implicit}} versus \textbf{\emph{explicit}} change. For example, an appointment may be postponed, which is an explicit change. On the other hand, the author may note in a follow-up statement that the appointment is in a sleep laboratory, which may cause us to re-evaluate for how long the original statement is valid, although the information itself has not changed.

The second dimension is a change to the \textbf{\emph{occurrence time}} versus the \textbf{\emph{duration}} of the information. For example, a flight may be delayed, in which case the occurrence time changes. Alternatively, the flight might have to be re-routed mid-air due to bad weather, in which case the duration changes. 

In our sample, we find that all four categories are present to a reasonable degree in both the \textsc{DEC} and \textsc{INC} classes. Generally, changes to the duration tend to be slightly more frequent than changes to the occurrence time. This makes sense, as the occurrence time is a dimension that is only present when the information occurs in the future, whereas the duration of temporal information can change irrespective of the occurrence time.

\section{Dataset}
We create a dataset for training and benchmarking \textsc{TVCP}, where each sample is a quintuple $<s_t, s_f, \textsc{TV}_d(s_t), \textsc{TV}_d^{s_f}(s_t), \textsc{TVCP}(s_t, s_f)>$. 

\begin{figure}[htbp]
    \centering
    \includegraphics[width=\columnwidth]{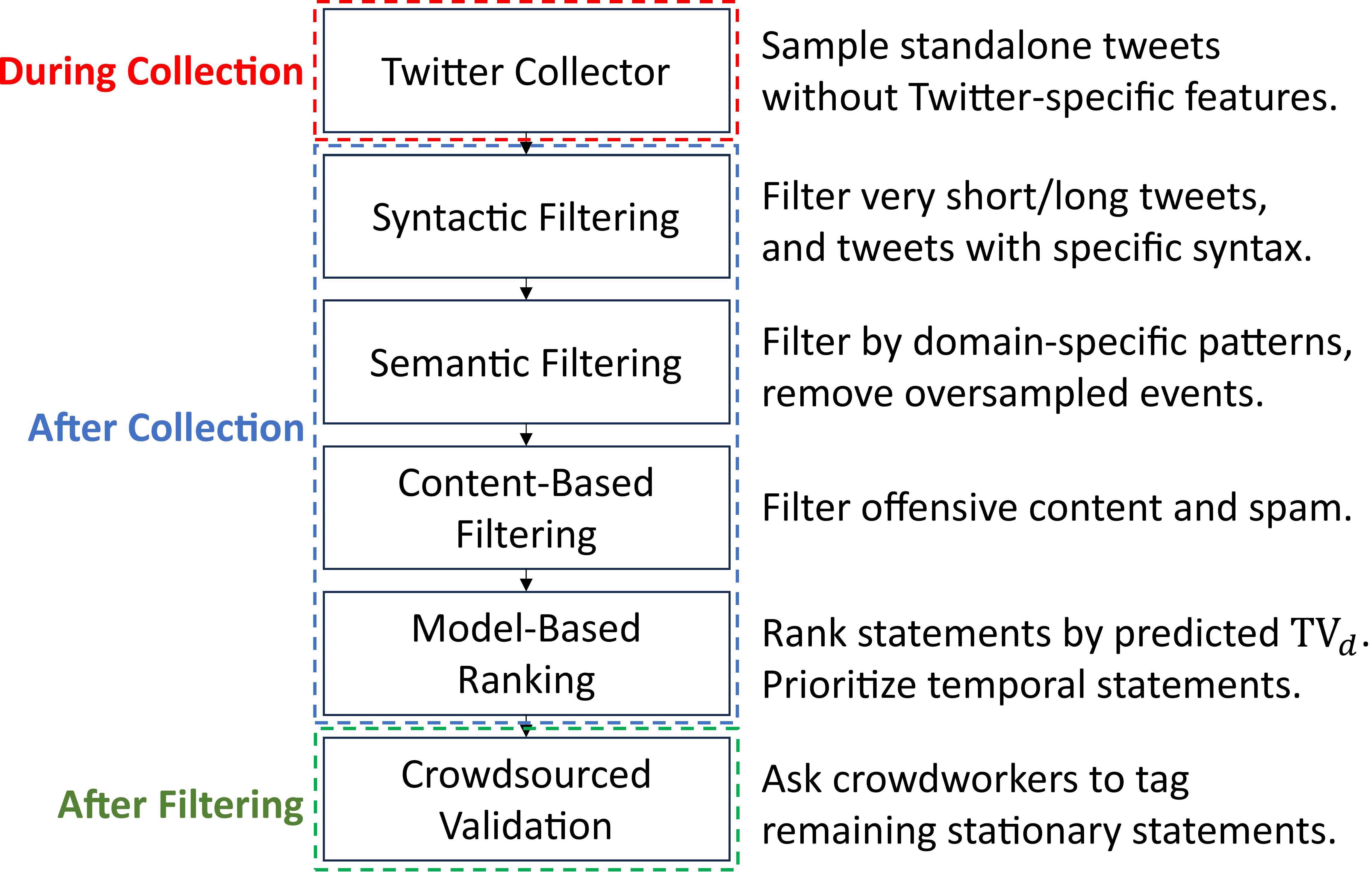}
    \caption{A summary of our tweet collection pipeline}
    \label{fig:twitterpipeline}
\end{figure}

$s_t$ is collected by querying the Twitter API for tweets with no external context (e.g., no tweets that are retweets or replies, or tweets containing media). We apply several pre-processing steps to remove tweets whose content may not be self-contained. We aim to minimize spam and offensive content by applying publicly available LMs and word-list-based filters. To decrease the number of stationary statements, we employ an ensemble classifier based on the \textsc{Almquist2019} \citep{content-expiry-date} and \textsc{CoTAK} datasets and select the most likely statements to contain temporal information. Finally, crowdworkers can tag any remaining stationary samples during the annotation process. A summary of our pre-processing pipeline is shown in Figure \ref{fig:twitterpipeline}. Our code, including all preprocessing steps, is published under the Apache 2.0 licence.

\begin{figure}[htbp]
    \centering
    \includegraphics[width=0.8\columnwidth]{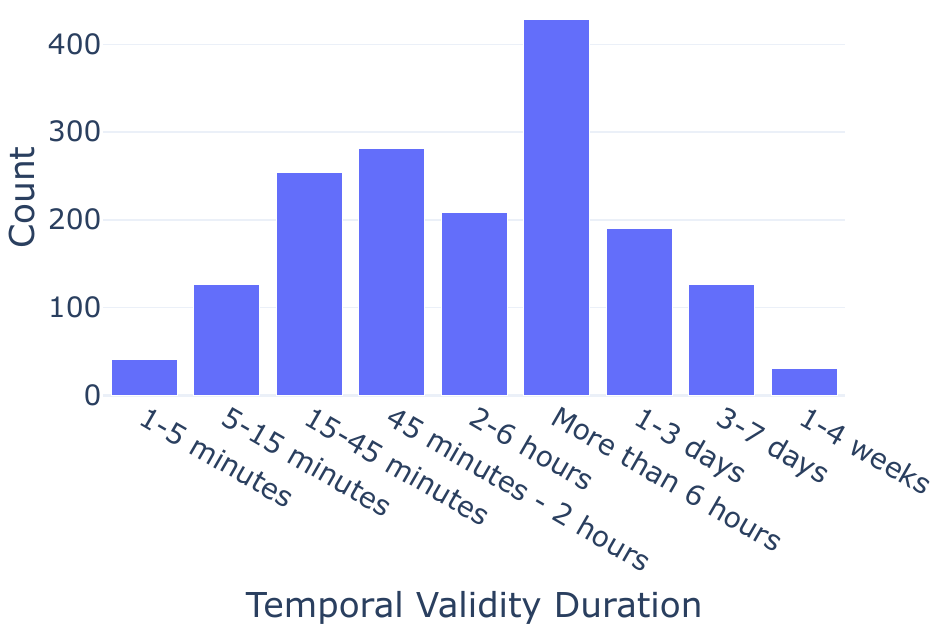}
    \caption{Distribution of \tvds labels (before temporal validity change) in our dataset}
    \label{fig:class_distribution}
\end{figure}

For each target statement, we ask two crowdworkers to estimate $\textsc{TV}_d(s_t)$ from the logarithmic class design shown in Equation \ref{eq:tvcpdef}, which is modelled after human timeline understanding \citep{logtimeline,logperception,logmemory}. If the annotators disagreed, we supplied a third vote. We discarded any tweets that were annotated as \emph{less than one minute}, \emph{more than one month}, or \emph{no time-sensitive information} (i.e., stationary), as well as tweets where no majority agreement could be reached. Of 2,996 annotated target tweets, 571 were discarded without a third annotation, 867 were added without a third annotation, 546 were discarded after providing a third vote, and 1,012 were added after providing a third vote. The distribution of resulting \tvds labels before temporal validity change is shown in Figure \ref{fig:class_distribution}.

\begin{equation}
\label{eq:tvcpdef}
\begin{split}
t \in \{<\textrm{1 minute}, \textrm{1-5 minutes}, \textrm{5-15 minutes},\\ \textrm{15-45 minutes}, \textrm{45 minutes-2 hours}, \textrm{2-6 hours},\\ \textrm{more than 6 hours},\textrm{1-3 days}, \textrm{3-7 days},\\\textrm{1-4 weeks}, \textrm{more than 1 month}\}\\
\end{split}
\end{equation}

\begin{figure}[htbp]
    \centering
    \includegraphics[width=\columnwidth]{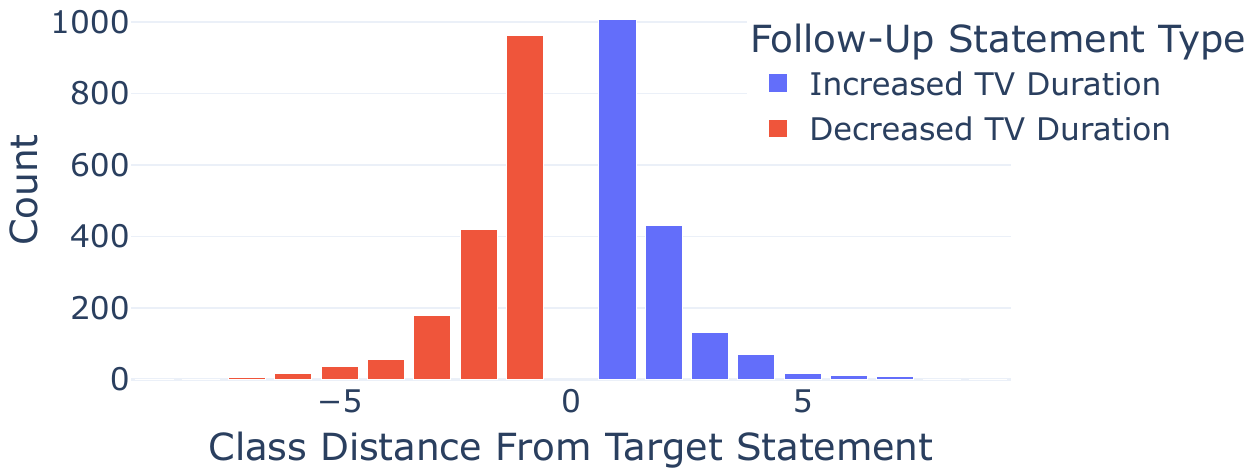}
    \caption{Temporal validity change delta distribution}
    \label{fig:changedelta}
\end{figure}

Both $s_f$ and $\textsc{TV}^{s_f}_d(s_t)$ were provided by a separate set of crowdworkers, given $s_t$ and $\textsc{TV}_d(s_t)$ as an input. In total, we collected 5,055 samples from 1,685 target statements. In Figure \ref{fig:changedelta}, we plot the \emph{temporal validity change delta}, which is the class distance between the original and the updated \tvds estimate. We find that, in most cases, the temporal validity duration of a target statement is shifted only by one class. 

All crowdsourcing tasks were set up on Amazon Mechanical Turk, using qualification tests, participation criteria, and manual verification of results to ensure high-quality samples (see Appendix \ref{sec:crowdsourcing-appendix}). We publish the resulting dataset for public use under the CC BY 4.0 licence. In accordance with the Twitter developer policy\footnote{\url{https://developer.twitter.com/en/developer-terms/policy}, accessed 12.10.2023}, we only publish the Tweet IDs of sourced statements. This also means original tweet authors retain the ability to delete their content, effectively removing it from the dataset.

\section{Experiments}
\label{sec:eval}
\subsection{Language Models}
We evaluate a set of transformer-based LMs on our dataset. We test four different archetypes in total:
\begin{itemize}
    \item \textsc{TransformerClassifier}: Builds a hidden representation from the sentence-embedding token of the concatenation of $s_t$ and $s_f$.
    
    \item \textsc{SiameseClassifier}: Builds a hidden representation from the concatenated embeddings  $[h_{s_t}, h_{s_f}, h_{s_t} - h_{s_f}, h_{s_t} \otimes h_{s_f}]$, where $h_{s_t}$ and $h_{s_f}$ are the sentence-embedding tokens of the target- and follow-up statement, respectively \citep{siamese,siamese-survey}.
    
    \item \textsc{SelfExplain} \citep{selfexplain}: Builds a hidden representation from the embeddings of spans between arbitrary tokens in either $s_t$ or $s_f$, selected by the model.
    
    \item \textsc{ChatGPT}: A chain-of-thought \citep{chainofthought} reasoning prompt based on few-shot learning (one sample per \textsc{TVCP} class), passed to the \lstinline!gpt-3.5-turbo! model via the OpenAI API. \footnote{This call uses the most recent \textsc{GPT3.5} model. We collected \textsc{ChatGPT} responses in July 2023.}
\end{itemize}

For the \textsc{TransformerClassifier} and \textsc{SiameseClassifier} pipelines, we evaluate \textsc{BERT-base-uncased} (\citealp{bert}; 110M parameters) and \textsc{RoBERTa-base} (\citealp{roberta}; 125M parameters) embeddings. For \textsc{SelfExplain}, we only test the original implementation with \textsc{RoBERTa-base} embeddings. To evaluate transfer learning from other TCS tasks, we test the \textsc{TransformerClassifier} pipeline on regular \textsc{BERT-base-uncased} pre-training weights as well as two variants \textsc{TacoLM} \citep{taco-lm} and \textsc{CoTAK} \citep{cta}, which have the same underlying architecture, but use weights fine-tuned on existing TCS datasets.

We use the \textsc{AdamW} optimizer \citep{adamw} with $\varepsilon=\textrm{1e-8}$, $\beta_1=0.9$, $\beta_2=0.999$, $\textrm{weight\_decay}=0.01$. We optimize for cross-entropy loss. \textsc{SelfExplain} adds an additional loss parameter in the form of squared span-attention weights, to encourage the model to more sharply choose which spans should be used to build the hidden representation.

We set the dropout probabilities and learning rates as defined in Table \ref{tab:hyperparameters} as a result of our hyperparameter optimization (see Appendix \ref{sec:hyperparam}). For all models, the hidden embedding size is 768. For some \textsc{RoBERTa}-based models, we freeze embedding layers (i.e., only fine-tune intermediate and classification weights), as training all parameters leads to poor performance.

\begin{table}[htbp]
\small
\centering
\begin{tabular}{|lrrl|}
\hline
Model & Dropout & LR & Frozen \\ \hline
TF - \textsc{BERT} & 0.25 & 1e-4 & False \\
S - \textsc{BERT} & 0.25 & 1e-4 & False \\
TF - \textsc{RoBERTa} & 0.25 & 1e-3 & True \\
S - \textsc{RoBERTa} & 0.10 & 1e-4 & True \\
\textsc{SelfExplain} & 0.00 & 2e-5 & False \\ \hline
\end{tabular}
\caption{Hyperparameter settings for different models. TF = \textsc{TransformerClassifier}, S = \textsc{SiameseClassifier}}
\label{tab:hyperparameters}
\end{table}

\subsection{Multitask Implementation}
For all archetypes except \textsc{ChatGPT}, we provide a second implementation, in which we add two regression layers that aim to respectively predict $\textsc{TV}_d(s_t)$ and $\textsc{TV}^{s_f}_d(s_t)$ from the same hidden representation. For these layers, we calculate the mean squared error between a single output neuron and a linear mapping of the \tvds class index to the range $[0,1]$. Our intuition is that embeddings with an understanding of \tvds may be better suited for $\textsc{TVCP}$. Inspiration for this approach are models that utilize the interplay between temporal dimensions to improve the TCS reasoning performance in LMs, such as \textsc{SymTime} \citep{tracie} or \textsc{SLEER} \citep{sleer}.

\subsection{Evaluation}
We evaluate two metrics, accuracy and \emph{exact match} (EM). Accuracy is simply the fraction of correctly classified samples. EM is the fraction of \emph{target statements} for which all three samples were correctly classified. This metric punishes inconsistency in the model more strictly, thus providing a better view of the true performance and task understanding of each model \citep{tcssurvey}, while disincentivizing shallow reasoning behaviours commonly seen in transformer models \citep{pitfalls,contemporarybias}.

We report the mean EM and accuracy across a five-fold cross-validation split. Each evaluation consists of 70\% training data, 10\% validation data, and 20\% test data. If the validation EM does not exceed the best previously observed value for 5 consecutive epochs, we stop training. The model with the best validation EM is used for evaluation on the test set. The results are shown in Table \ref{tab:eval}.

\begin{table}[htbp]
    \small
    \centering
    \begin{tabular}{|lrr|}
         \hline
         \rule{0pt}{2.5ex}
         Model & $\overline{\textrm{Acc}}$ (+ MT) & $\overline{\textrm{EM}}$ (+ MT)\\
         \hline
         TF - \textsc{RoBERTa} & $64.0$ $(+ 1.5)$ & $21.2$ $(+ 2.5)$\\
         \textsc{ChatGPT} & $66.3\;\;$ (N/A) & $29.3\;\;$ (N/A)\\
         S - \textsc{RoBERTa} & $78.7$ $(+ 1.1)$ & $48.2$ $(+ 2.1)$\\
         TF - \textsc{CoTAK} & $83.2$ $ (+ 0.6)$ & $58.2$ $(+ 1.4)$\\
         S - \textsc{BERT} & $83.8$ $(- 0.3)$ & $59.1$ $(- 1.5)$\\
         TF - \textsc{TacoLM} & $83.5$ $(+ 1.4)$ & $59.1$ $(+ 2.9)$\\
         TF - \textsc{BERT} & $84.8$ $(- 0.2)$ & $61.2$ $(+ 0.9)$\\
         \textsc{SelfExplain} & $88.5$ $(+ 1.1)$ & $69.8$ $(+ 2.8)$\\
         \hline
    \end{tabular}
    \caption{Model evaluation results, sorted by mean EM score. TF = \textsc{TransformerClassifier}, S = \textsc{SiameseClassifier}, MT = Multitask Implementation}
    \label{tab:eval}
\end{table}

We note a positive impact on the EM score from implementing multitasking in all models except the Siamese architecture with \textsc{BERT}-based embeddings. We use bootstrapping to calculate the statistical significance of implementing multitask learning on the best-performing model, \textsc{SelfExplain}, evaluating the number of bootstrap samples in which the multitask implementation outperforms regular \textsc{SelfExplain}. We find $p=0.0027$ for accuracy, with a 95\% confidence interval of $[0.0036, 0.0192]$. For EM, $p=0.0025$, with a 95\% confidence interval of $[0.0089, 0.0487]$.

The use of weights from other TCS tasks does not seem to have a positive impact on the performance of the \textsc{TransformerClassifier} pipeline. It is possible that, although the resulting embeddings are more aligned with temporal properties \citep{taco-lm}, other important information in the embeddings is lost, leading to an overall decreased performance.

Due to some \textsc{RoBERTa}-based models having frozen embedding layers, the baseline performance by \textsc{RoBERTa} is much worse, but it improves much more when switching to the \textsc{SiameseClassifier} implementation. We hypothesize that \textsc{RoBERTa}'s sentence embedding token, \lstinline!<s>!, may contain less information about the full sequence than \textsc{BERT}'s \lstinline![SEP]! token due to the lack of a next-sentence-prediction task during pre-training.

\textsc{ChatGPT} ranks among the lower-performing models, which is consistent with other studies on TCS understanding \citep{chatgpt-tcs}. Its shortcomings may be due to the few-shot learning approach and a lack of knowledge about dataset specifics traits, which a trained classifier could leverage. Additionally, we do not specify our class design in the \textsc{ChatGPT} prompt, which could make it harder for \textsc{ChatGPT} to isolate the \textsc{UNC} class. 

\begin{figure}[htbp]
    \centering
    \includegraphics[width=\columnwidth]{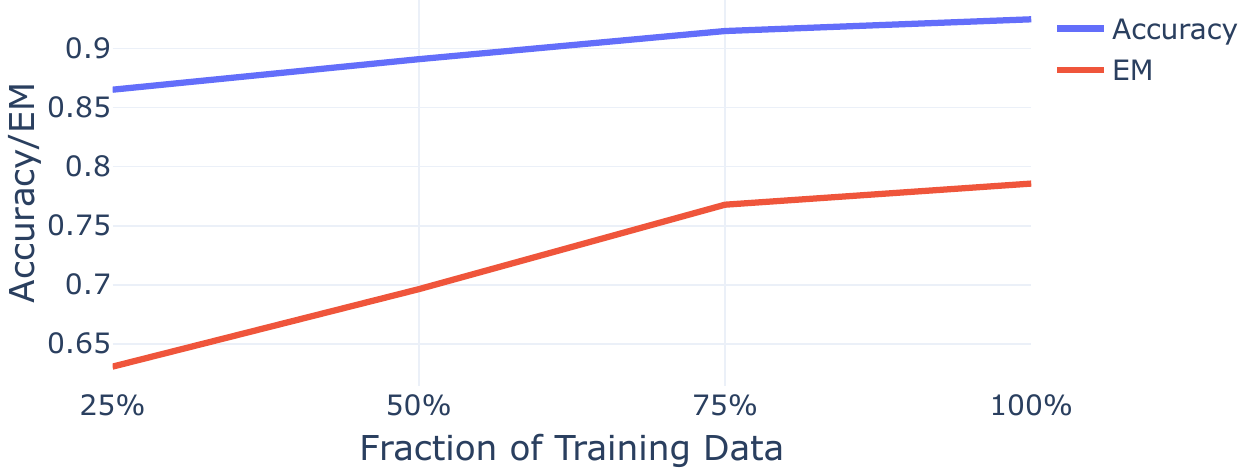}
    \caption{Training data vs. performance metrics in \textsc{MultiTask}}
    \label{fig:data-vs-em}
\end{figure}

To evaluate the impact of training data quantity on classifier performance, we train our best-performing classifier (\textsc{SelfExplain} with multitasking, which we dub \textsc{MultiTask}) on a single train-val-test split (80\%/10\%/10\%) with different amounts of training data. The results can be seen in Figure \ref{fig:data-vs-em}. We find that performance increases as more data is used for training, but this effect starts to diminish as we approach 100\% of our training data.

\begin{figure}[htbp]
    \centering
    \includegraphics[width=\columnwidth]{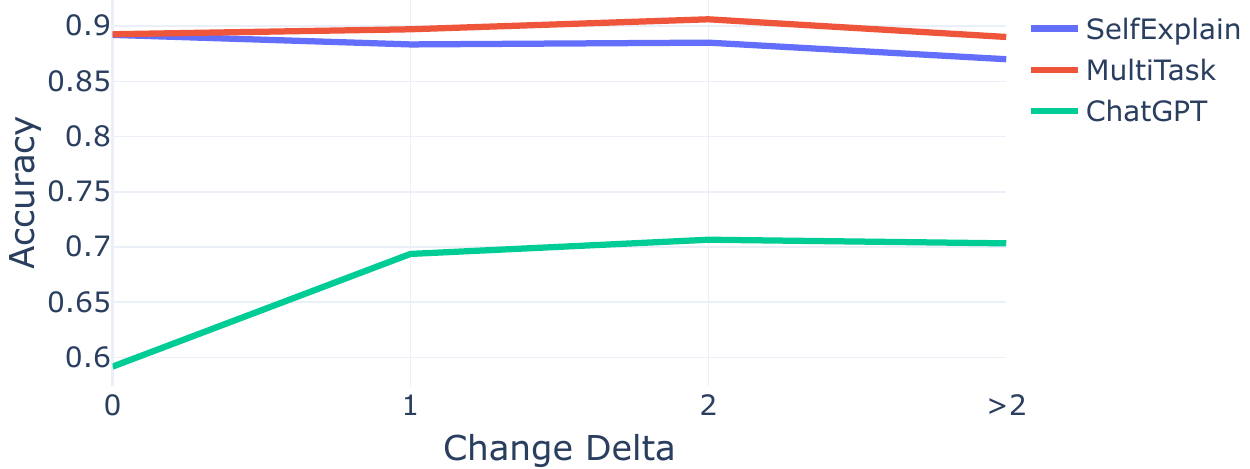}
    \caption{Temporal validity change delta vs. accuracy in \textsc{MultiTask}, \textsc{SelfExplain} and \textsc{ChatGPT}}
    \label{fig:delta-vs-em}
\end{figure}

In testing \textsc{SelfExplain} and \textsc{MultiTask} on various temporal validity change deltas (Figure \ref{fig:delta-vs-em}), we find they perform comparably on the \textsc{UNC} class, but \textsc{MultiTask} slightly outperforms \textsc{SelfExplain} on all delta values greater than zero. While \textsc{ChatGPT}'s subpar performance in the \textsc{UNC} class can partially be attributed to prompt design, it continues to lag far behind other models in the \textsc{DEC} and \textsc{INC} classes.

All models were trained and evaluated on an MSI GeForce RTX 3080 GAMING X TRIO 10G GPU using CUDA 11.7. Training and evaluation of all final models as well as hyperparameter tests took around 15 GPU hours.

\section{Conclusion and Future Work}
In this work, we have introduced \textsc{TVCP}, an NLP task, to aid in the accurate determination of the temporal validity duration of text by incorporating surrounding context. We create a benchmark dataset for our task and provide a set of baseline evaluation results for our dataset. We find that the performance of most classifiers can be improved by explicitly incorporating the temporal validity duration as a loss signal during training to improve the resulting embeddings. Despite the impressive feats performed by foundation models, we report, similar to previous work \citep{chatgpt-tcs}, poor performance in the TCS domain. These findings show that users should carefully evaluate whether a model like \textsc{ChatGPT} properly understands a given task before choosing it over smaller, fine-tuned LMs. We hypothesize that the performance of all models could further increase with additional training data.

Possible future work includes using the provided dataset and classifiers to collect a larger number of \textsc{TVCP} samples and annotating them with an updated temporal validity duration. A comparison of context-aware \tvds classifiers with prior models, like those by \citet{content-expiry-date}, would shed light on the significance of accurate semantic segmentation between target and context. Similarly, the use of our dataset for generative approaches could be explored, for example, in the context of generative adversarial networks. For our multitasking implementation, directions for future work could be changes to hyperparameters such as the weight of the auxiliary loss, changes to the definition of the auxiliary task (e.g., log-scaled regression or ordinal classification), or even entirely new auxiliary tasks. Finally, current methods face limitations due to the effort of manual removal of stationary samples (\citealp{content-expiry-date}; ours) or altering task definitions to avoid them \citep{tnli,cta}. Research into models differentiating temporal and stationary information could enhance the development and definition of future TCS reasoning tasks.

\section*{Limitations}
Although we focus on creating a reproducible training- and evaluation environment, some variables are out of our control. For example, bit-wise reproducibility is only guaranteed on the same CUDA toolkit version and when executed on a GPU with the same architecture and the same number of streaming multiprocessors. This means that an exact reproduction of the models discussed in this article may not be possible. Nevertheless, we expect trends to remain the same across GPU architectures.

The use of \textsc{ChatGPT} as an example of foundation model performance may be limiting due to its black box design. In the future, open-source models, such as \textsc{Llama 2} \citep{llama}, could be evaluated to improve the reproducibility of foundation model performance claims. We chose to benchmark \textsc{ChatGPT} due to its common use as a baseline and in end-user scenarios, but the evaluation results may not be transferrable to other foundation models or even other versions of \textsc{ChatGPT}.

One of the major limitations of our approach is likely the dataset size. Although a relatively small dataset size is common in TCS reasoning, we find that our model performance still increases with the amount of training data used. The existing synthesized context statements in our dataset could be used to bootstrap an approach for automatically extracting additional samples from social media to alleviate this issue.

The data we collect is not personal in nature. However, the possibility of latent demographic biases in our data exists, for example, with respect to certain language structures or expressions used in the creation of follow-up statements. This could lead to the propagation of any such bias when the dataset is used to bootstrap further data collection, which should be considered in future work.

Our external validity is mainly threatened by two factors. First, our context statements are crowdsourced. While we apply several steps to ensure the produced context is sensible, it is unclear whether downstream context, such as on social media platforms, manifests in similar structures as in our dataset, with respect to traits such as sentence length, grammaticality, and phrasing.

Second, similar to how pre-training weights from other TCS tasks do not seem to improve the classifier performance on our dataset, the weights generated as part of our training process are likely very task-specific, and may not generalize well to other tasks or text sources.

Overall, we recommend the use of the \textsc{TVCP} dataset and classifiers for bootstrapping further research into combining the duration- and inference-based temporal validity tasks, as well as research into directly predicting updated temporal validity durations and improving the generalizability to different text sources, rather than for a direct downstream task application.

\bibliography{custom}
\bibliographystyle{acl_natbib}

\appendix
\section{Crowdsourcing Definitions}
\label{sec:crowdsourcing-appendix}
In this section, we provide details on the crowdsourcing implementation. As noted, we use Amazon Mechanical Turk to collect crowdsourced data from participants. 

\subsection{Temporal Validity Duration Estimation}
We assume the average layman is not familiar with the term \emph{temporal validity}. Thus, we define the task as \q{determining how long the information within the tweet remains relevant after its publication}, i.e., for how long the user would consider the tweet timely and relevant. We provide the option \emph{no time-sensitive information}, which should be selected when tweets do not contain any information, when information is not expected to change over time, or when it is fully contained in the past.

\begin{figure*}[htbp]
    \centering
    \includegraphics[width=0.8\textwidth]{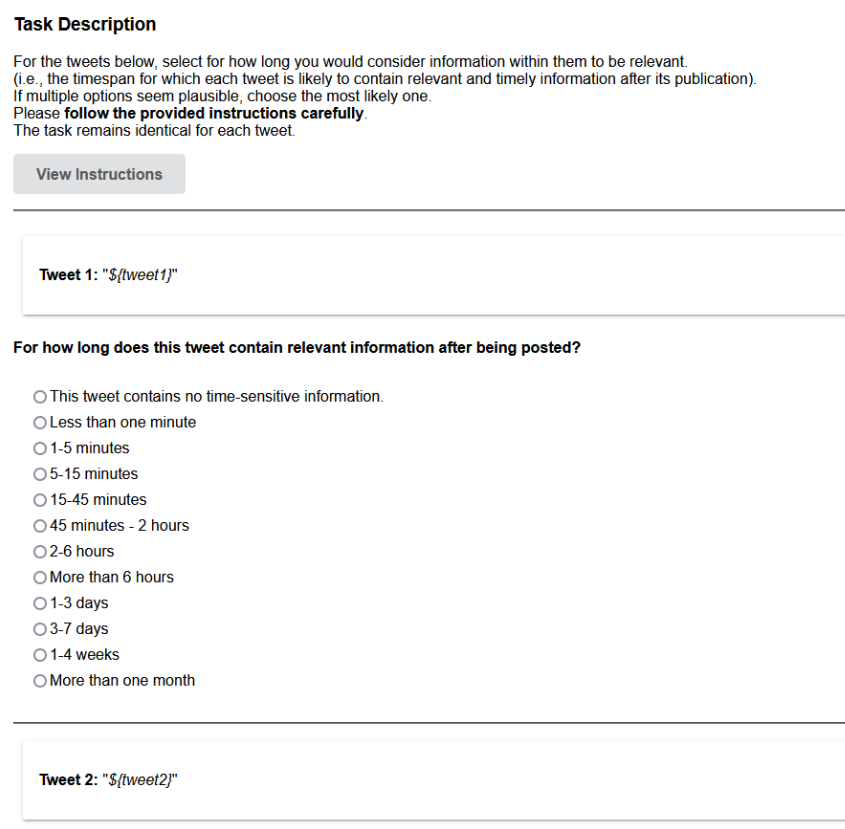}
    \caption{The interface of the temporal validity duration estimation task}
    \label{fig:tvtask-appendix}
\end{figure*}

\begin{figure*}[htbp]
    \centering
    \includegraphics[width=0.8\textwidth]{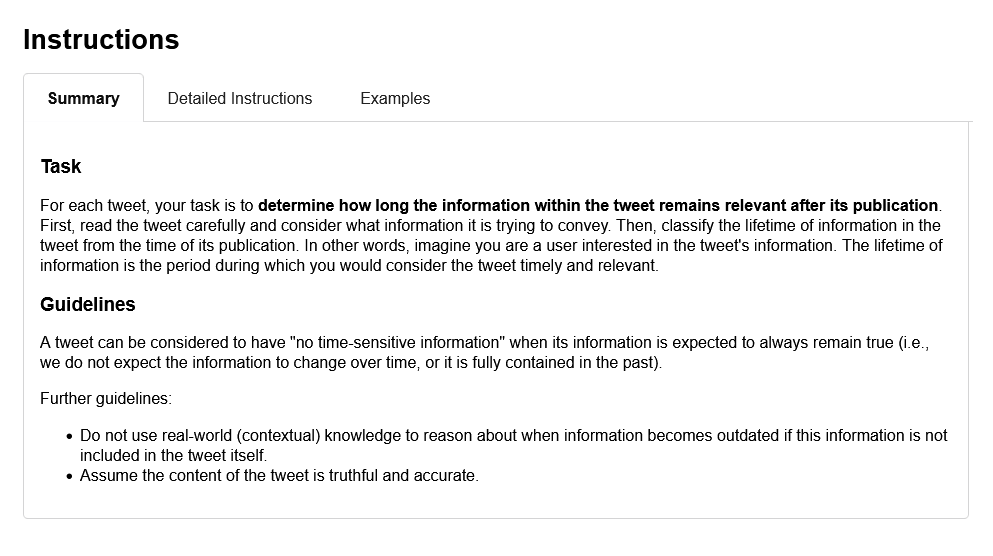}
    \caption{The summary section of the temporal validity duration estimation task guidelines}
    \label{fig:tvsummary-appendix}
\end{figure*}

\begin{figure*}[htbp]
    \centering
    \includegraphics[width=0.8\textwidth]{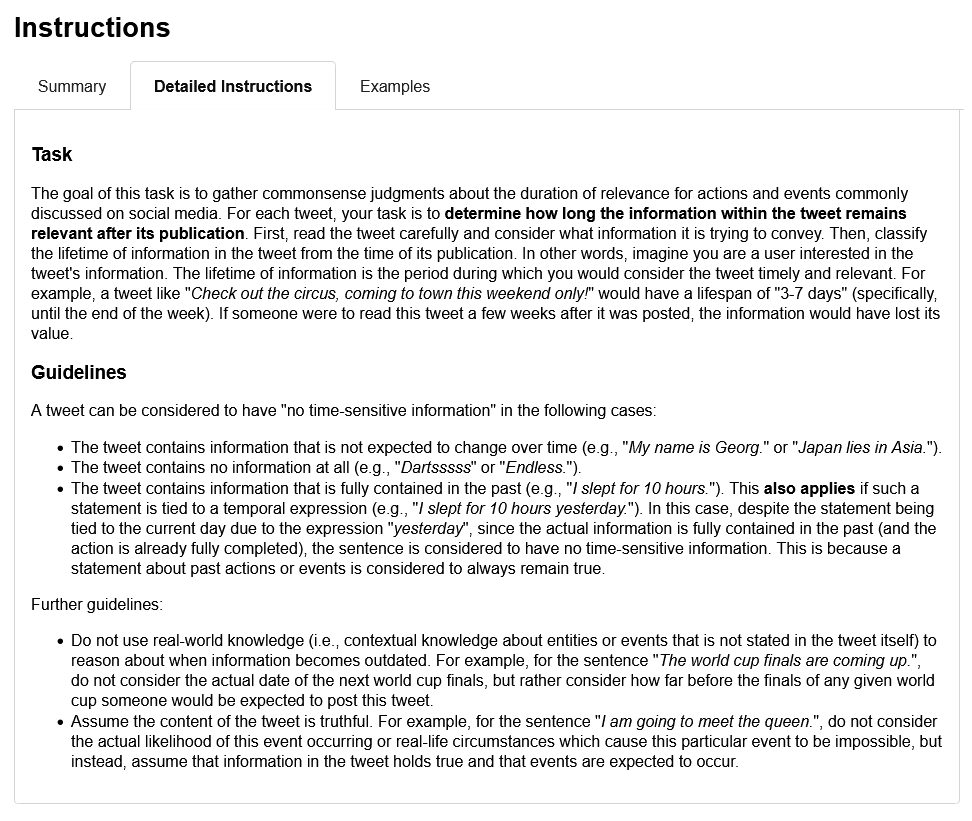}
    \caption{The detailed description of the temporal validity duration estimation task guidelines}
    \label{fig:tvdetailed-appendix}
\end{figure*}

\begin{figure*}[htbp]
    \centering
    \includegraphics[width=0.8\textwidth]{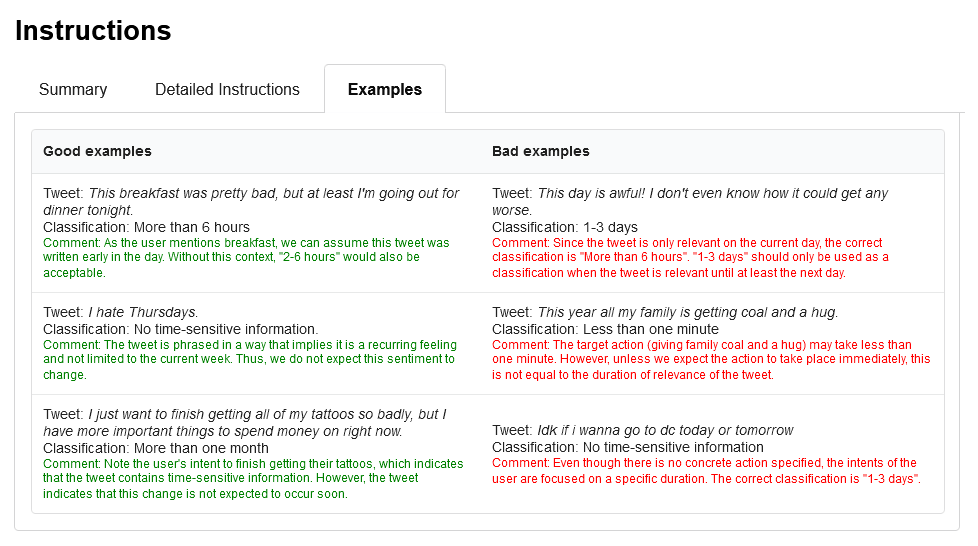}
    \caption{The examples section of the temporal validity duration estimation task guidelines}
    \label{fig:tvexamples-appendix}
\end{figure*}

The task is otherwise a relatively straightforward classification task. We split our dataset into batches of 10 samples that are grouped into a single \emph{human intelligence task} (HIT). For each HIT, we offer a compensation of $\textrm{USD}0.25$, based on an estimated 6-9 seconds of processing time per individual statement (i.e., 60-90 seconds per HIT). Figures \ref{fig:tvtask-appendix} to \ref{fig:tvexamples-appendix} show the crowdsourcing layout.

\subsection{Follow-Up Content Generation}
\begin{figure*}[htbp]
    \centering
    \includegraphics[width=\textwidth]{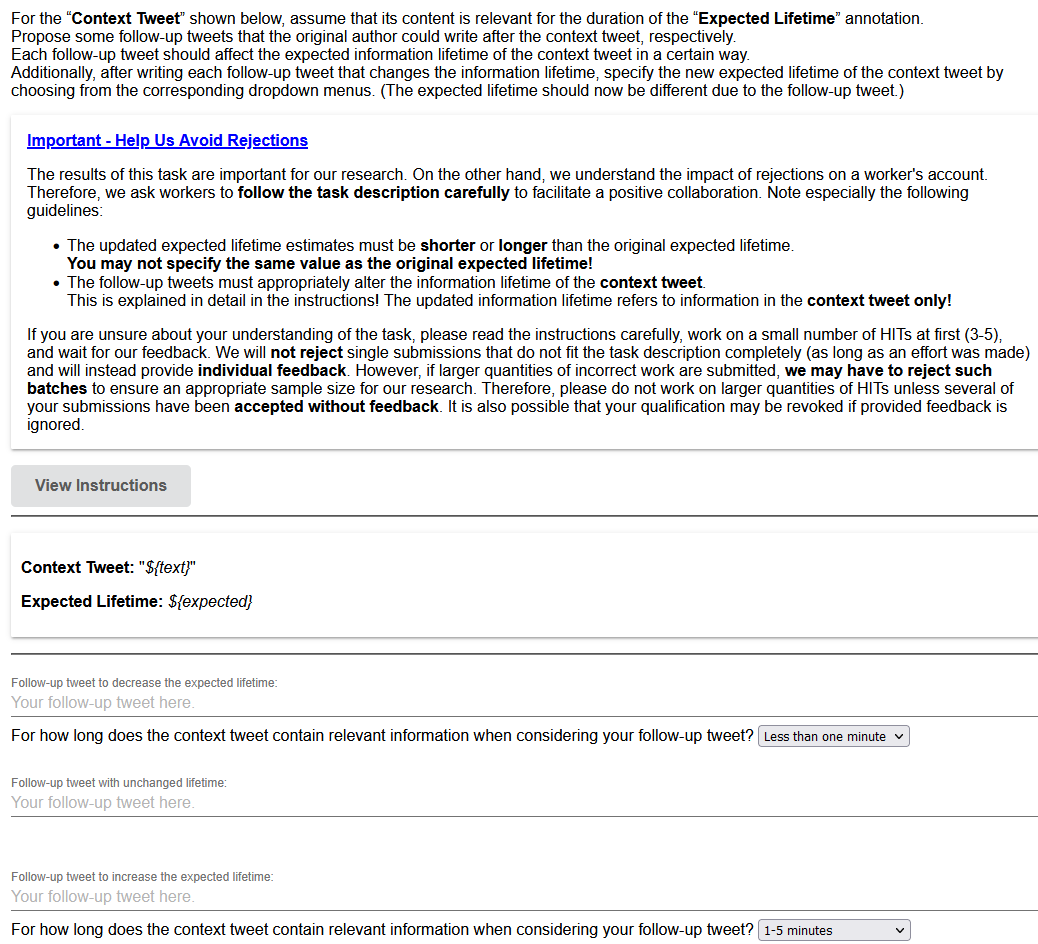}
    \caption{The interface of the follow-up content generation task}
    \label{fig:fgtask-appendix}
\end{figure*}

\begin{figure*}[htbp]
    \centering
    \includegraphics[width=0.8\textwidth]{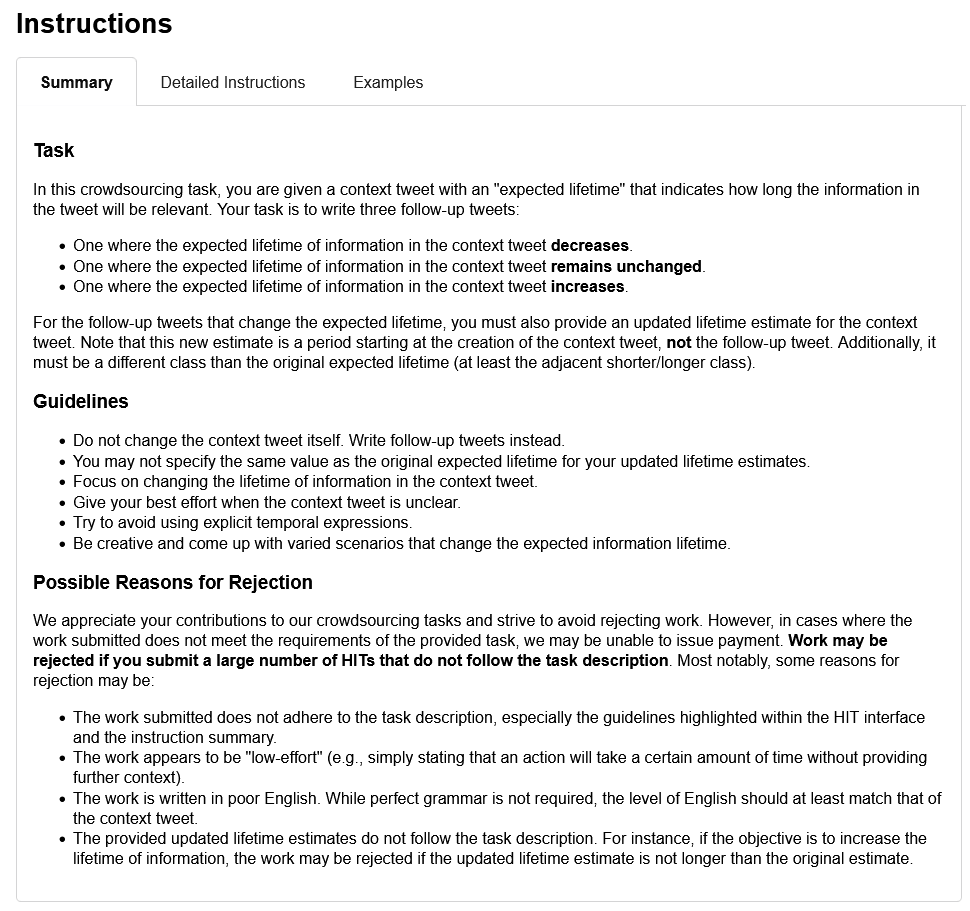}
    \caption{The summary section of the follow-up content generation task guidelines}
    \label{fig:fgsummary-appendix}
\end{figure*}

\begin{figure*}[htbp]
    \centering
    \includegraphics[width=0.8\textwidth]{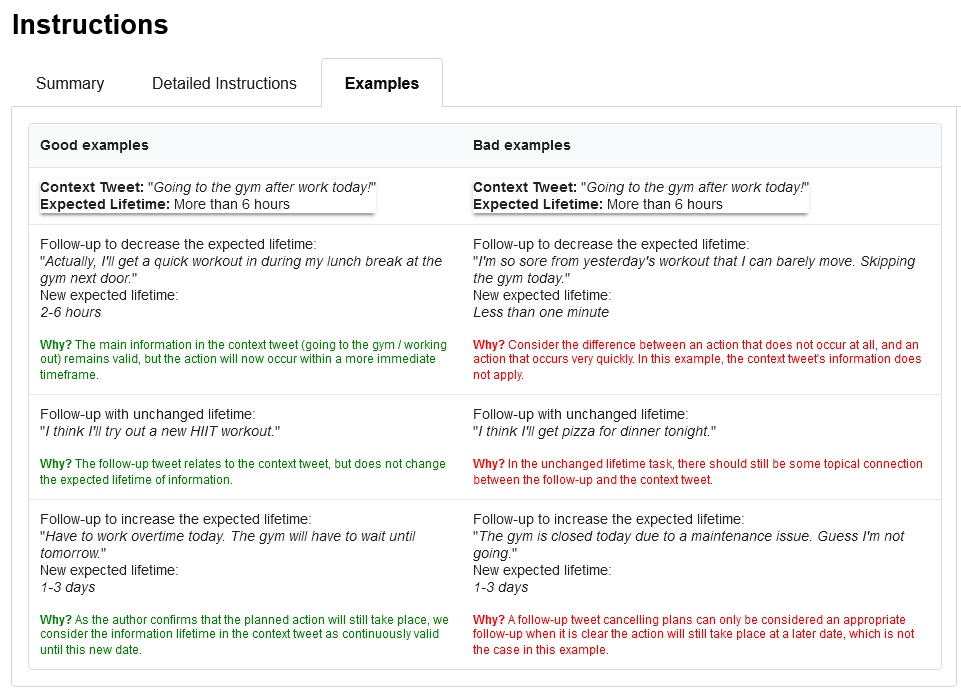}
    \caption{The examples section of the follow-up content generation task guidelines}
    \label{fig:fgexamples-appendix}
\end{figure*}

Compared to the temporal validity duration estimation task, the follow-up content generation task requires a much more robust understanding of the overall concept of temporal validity and the respective semantic roles of the target- and follow-up statements. Hence, we focus on providing a more detailed explanation of the task. Figures \ref{fig:fgtask-appendix} to \ref{fig:fgexamples-appendix} show the crowdsourcing setup. The detailed instructions tab is not listed due to its length, but contains instructions that can also be found in the public repository.

Notably, we labelled the target statement as \emph{context tweet} rather than \emph{target tweet} in this crowdsourcing task to emphasize that participants should not alter this statement directly, as this was a problem that occurred somewhat frequently during pilot tests. This contrasts with our formal definition of \textsc{TVCP}, where providing context is the role of the follow-up statement.

Each HIT requires participants to provide three follow-up statements, one for each \textsc{TVCP} class (\textsc{DEC}, \textsc{UNC}, \textsc{INC}). For each HIT, we offer a compensation of $\textrm{USD}0.35$. We base our compensation on an estimated 30–40 seconds of processing time per follow-up statement (i.e., 90–120 seconds per HIT) due to the creative writing involved .

\subsection{Discouraging Dishonest Activity}
\label{subsec:appendix-dishonest}
In initial pilot runs, we find that many submissions are the result of spam, dishonest activity, or a complete lack of task understanding, with many provided annotations being inexplicable by common sense in any possible interpretation of the statement. 

To increase the quality of work on both tasks, we introduced three measures. First, we required participants to have an overall approval rate of 90\% on the platform, as well as 1,000 approved HITs. Without these requirements, the amount of blatant spam (e.g., copy-pasted content) increases significantly. 

Second, we devised qualification tests for both tasks. Participants had to determine the temporal validity durations for a set of sample statements to work on the temporal validity duration estimation task, and determine the correctness of follow-up statements and their updated duration labels to work on the follow-up content generation task.

Finally, we vet all participants' responses individually up to a certain threshold. For each task, we manually verify the first 20 submissions of each annotator on their quality. We provide feedback and manually adapt submissions when they are partially incorrect. If submission quality is appropriate by the time a participant reaches 20 submitted HITs, we consider them as trusted, and only spot-check every 5th submission thereafter. If submission quality does not sufficiently improve at this point, we prohibit the participant from further working on the task. 

Despite these efforts, the follow-up content generation task specifically still received several low-quality submissions that had to be manually filtered out and corrected. In future work, a preferable approach may be to replace the qualification test with an unpaid qualification HIT, in which a feedback loop between participants and requesters can be established on data that will not be included in the final dataset, and participants can manually be assigned a qualification once their quality of work is sufficient.

\section{ChatGPT Setup}
\label{sec:chatgpt-appendix}
We provide the following system prompt to the \textsc{ChatGPT} API:
\begin{quote}
    \q{You are a language model specializing in temporal commonsense reasoning. Each prompt contains Sentence A and Sentence B. You should determine whether Sentence B changes the expected temporal validity duration of Sentence A, i.e., the duration for which the information in Sentence A is expected to be relevant to a reader.

    To achieve this, in your responses, first, estimate for how long the average reader may expect Sentence A to be relevant on its own. Then, consider if the information introduced in Sentence B increases or decreases this duration. Surround this explanation in triple backticks (\lstinline!```!).

    After your explanation, respond with one of the three possible classes corresponding to your explanation: Decreased, Neutral, or Increased.}
\end{quote}

After this system prompt, we provide three sample responses, one for each of the classes. These sample responses are shown in Figure \ref{fig:chatgpt-examples}. 

\begin{figure}[t!]
    \noindent\rule[1pt]{0.99\columnwidth}{1pt}

    \noindent\textbf{Sentence A}: I'm ready to go to the beach\\
    \textbf{Sentence B}: I forgot all the beach towels are still in the dryer, but I'll be ready to go as soon as the dryer's done running.\\
    \textbf{Target Class}: Increased\\
    \textbf{Sample Explanation}: Going to the beach may take a few minutes to an hour, depending on the distance. However, if the author first needs to wait on the dryer to finish in order to retrieve their beach towels, this may take an additional 30-60 minutes.
    
    \noindent\rule[1pt]{0.99\columnwidth}{1pt}
    \noindent\rule[1pt]{0.99\columnwidth}{1pt}

    \noindent\textbf{Sentence A}: taking bad thoughts out of my mind thru grinding my assignments\\
    \textbf{Sentence B}: I just have to get through a short math homework assignment and memorize a few spelling words so it shouldn't take long.\\
    \textbf{Target Class}: Decreased\\
    \textbf{Sample Explanation}: Grinding through assignments may take several hours, depending on the number of assignments to complete. In Sentence B, the author states they only have a few short assignments remaining, so they may only take an hour or less to finish them.
    
    \noindent\rule[1pt]{0.99\columnwidth}{1pt}
    \noindent\rule[1pt]{0.99\columnwidth}{1pt}

    \noindent\textbf{Sentence A}: Slide to my dm guys, come on\\
    \textbf{Sentence B}: Instagram DMs are such a fun way to communicate.\\
    \textbf{Target Class}: Neutral\\
    \textbf{Sample Explanation}: The author encourages people to direct message them, which may be relevant for several minutes to a few hours. Sentence B does not change the duration for which Sentence A is expected to be relevant.
    
    \noindent\rule[1pt]{0.99\columnwidth}{1pt}
\caption{Sample items, target classes, and explanations provided to \textsc{ChatGPT} for few-shot reasoning}
\label{fig:chatgpt-examples}
\end{figure}

Similar to the crowdsourcing task setup, we use the concept of the \emph{expected relevance duration} of the target statement (called Statement A in the \textsc{ChatGPT} prompt) to explain statement-level temporal validity. Additionally, instead of prompting the model to classify the sample directly, we ask it to provide an explanation for its decision based on chain-of-thought reasoning. \citet{chainofthought} show that chain-of-thought prompting significantly increases several types of reasoning capabilities, including commonsense, in LLMs. We prompt \textsc{ChatGPT} to first estimate the temporal validity duration of the target statement on its own. In a second step, the model should then determine if the information introduced in the follow-up statement changes this temporal validity duration. After giving its explanation, the model should respond with one of the three target classes.

\section{Hyperparameters}
\label{sec:hyperparam}
We perform hyperparameter testing regarding dropout probability before the classification layer ($0.1$, $0.25$, $0.5$), the base learning rate (1e-2, 1e-3, 1e-4), and whether to freeze embedding layers (i.e., training only intermediary and classification layers). For both \textsc{BERT} and \textsc{RoBERTa} in the frozen and unfrozen setting, we perform grid-search over the learning rate and dropout probability. For these benchmarks, we use a predefined train-val-test split (80\%/10\%/10\%). The remaining setup is the same as in Section \ref{sec:eval}.

Table \ref{tab:hyperparam-results} shows the three best-performing configurations for \textsc{BERT} and \textsc{RoBERTa} in the freeze and nofreeze settings, respectively, on the \textsc{TransformerClassifier} pipeline. Table \ref{tab:hyperparam-results-siamese} shows the same results for the \textsc{SiameseClassifier} pipeline. 

\begin{table}[htbp]
\small
    \centering
    \begin{tabular}{|lrrrr|}
    \hline
     Model & DO & LR & \#Epochs & EM\\
    \hline
     \textsc{BERT}-nofreeze & 0.25 & 1e-4 & 5 & 0.613\\
     \textsc{BERT}-nofreeze & 0.10 & 1e-4 & 6 & 0.548\\
     \textsc{BERT}-nofreeze & 0.50 & 1e-4 & 4 & 0.548\\
     \textsc{BERT} & 0.25 & 1e-4 & 17 & 0.321\\
     \textsc{BERT} & 0.10 & 1e-4 & 8 & 0.315\\
     \textsc{BERT} & 0.10 & 1e-3 & 10 & 0.304\\
     \textsc{RoBERTa} & 0.25 & 1e-3 & 14 & 0.262\\
     \textsc{RoBERTa} & 0.10 & 1e-4 & 16 & 0.256\\
     \textsc{RoBERTa} & 0.50 & 1e-3 & 15 & 0.238\\
     \textsc{RoBERTa}-nofreeze & 0.25 & 1e-3 & 1 & 0.000\\
     \textsc{RoBERTa}-nofreeze & 0.50 & 1e-3 & 1 & 0.000\\
     \textsc{RoBERTa}-nofreeze & 0.10 & 1e-4 & 1 & 0.000\\
    \hline
    \end{tabular}
    \caption{Best three models for each of the proposed configurations in the \textsc{TransformerClassifier} pipeline}
    \label{tab:hyperparam-results}
\end{table}

\begin{table}[htbp]
\small
    \centering
    \begin{tabular}{|lrrrr|}
    \hline
     Model & DO & LR & \#Epoch & EM\\
    \hline
         \textsc{BERT}-nofreeze & 0.25 & 1e-4 & 7 & 0.589\\
         \textsc{BERT}-nofreeze & 0.10 & 1e-4 & 4 & 0.577\\
         \textsc{BERT}-nofreeze & 0.50 & 1e-4 & 2 & 0.565\\
         \textsc{RoBERTa} & 0.10 & 1e-4 & 21 & 0.548\\
         \textsc{RoBERTa} & 0.50 & 1e-4 & 13 & 0.518\\
         \textsc{RoBERTa} & 0.25 & 1e-4 & 17 & 0.512\\
         \textsc{BERT} & 0.50 & 1e-4 & 9 & 0.387\\
         \textsc{BERT} & 0.25 & 1e-3 & 8 & 0.357\\
         \textsc{BERT} & 0.25 & 1e-4 & 5 & 0.339\\
         \textsc{RoBERTa}-nofreeze & 0.25 & 1e-3 & 1 & 0.000\\
         \textsc{RoBERTa}-nofreeze & 0.50 & 1e-3 & 1 & 0.000\\
         \textsc{RoBERTa}-nofreeze & 0.10 & 1e-4 & 1 & 0.000\\
         \hline
    \end{tabular}
    \caption{Best three models for each of the proposed configurations in the \textsc{SiameseClassifier} pipeline}
    \label{tab:hyperparam-results-siamese}
\end{table}

The most notable finding appears to be that \textsc{RoBERTa} gets stuck in a false minimum of predicting a constant class when embedding layers are unfrozen, leading to an accuracy of 0.33 and an EM of 0. Hence, we freeze embedding layers for these model types in our main evaluation. As noted in Section \ref{sec:eval}, a possible reason for this behaviour could be differences in the embeddings contained within \textsc{BERT}'s \lstinline![CLS]! and \textsc{RoBERTa}'s \lstinline!<s>! token.

\end{document}